# A deep learning and machine learning approach to predict neonatal death in the context of São Paulo

**Mohon Raihan[1], Plabon Kumar Saha[1], Rajan Das Gupta[2], A Z M Tahmidul Kabir[3], Afia Anjum Tamanna[4], Md. Harun-Ur-Rashid[5], Adnan Bin Abdus Salam[1], Md Tanvir Anjum[6], A Z M Ahteshamul Kabir[7]**

[1]Department of Computer Science and Engineering, Faculty of Science and Technology, American International University-Bangladesh, Dhaka, Bangladesh
[2]Department of Computer Science, Faculty of Mathematical & Physical Sciences, Jahangirnagar University, Dhaka, Bangladesh
[3]Department of Electrical and Electronic Engineering, Faculty of Engineering, American International University-Bangladesh, Dhaka, Bangladesh
[4]Department of Computer Science and Engineering, Faculty of Engineering and Technology, University of Dhaka, Dhaka, Bangladesh
[5]Department of Computer Science and Engineering, Faculty of Science & Engineering, United International University-Bangladesh, Dhaka, Bangladesh
[6]Department of Computer Science and Software Engineering, Faculty of Science and Technology, American International University-Bangladesh, Dhaka, Bangladesh
[7]Department of Predictive Analytics, Faculty of Science and Engineering, Curtin University, Perth, Australia

**ABSTRACT**

Neonatal death is still a concerning reality for underdeveloped and even for some of the developed countries. Worldwide data indicate that 26.693 babies out of 1,000 births according to Macro Trades. To reduce the death early prediction of endangered baby is crucial. An early prediction enables the opportunity to take ample care of the child and mother so that an early child death can be avoided. Machine learning was used to figure out whether a newborn baby is at risk. To train the predictive model historical data of 1.4 million newborn child data was used. Machine learning and deep learning techniques such as Logical regression, K nearest neighbor, Random Forest classifier, Extreme gradient boosting (XGboost), convolutional neural network, long short-term memory (LSTM). were implemented using the dataset to find out the most robust model which model is the most accurate to identify the mortality of a newborn. From all the machine learning algorithms, the XGboost and random classifier had the best accuracy with 94%, and from the deep learning model, the LSTM had the best outcome with 99% accuracy. Thus, using LSTM of the model shall be best suited to predict whether precaution for a child is necessary.

*Corresponding Author:*

A Z M Tahmidul Kabir
Department of Electrical and Electronic Engineering, American International University-Bangladesh
Dhaka-1229, Bangladesh
Email: tahmidulkabir@gmail.com

## 1. INTRODUCTION

The infant mortality rate (IMR) is the number of children under the age of one who dies for every 1,000 children. Neonatal death is defined as the death of a newborn within the first 28 days (approximately four weeks) after birth. Lack of first aid during pregnancy and after birth is usually blamed for neonatal deaths. In 2022, the current infant mortality rate for the World is 26,693 [1]. From the data provided by the world bank [2], it is evident that the mortality rate is extremely high in some places. It especially happens in the least developed countries like Chad, Nigeria, and Sierra Leone [2]. Infant mortality (IM) is a key indicator of a population's overall health, as well as an approximate predictor of social inequality and financial inequality position. It also shows the availability and quality of healthcare and medical technologies in a certain location. The IMR is used to assess needs and evaluate the efficacy of public interventions.

Due to the unlimited access to data through various sources in modern times, data-driven decision-making has become popular and optimal. Data-driven decisions are being used in various fields such as medical science, sports, autonomous vehicles, education, and business. A popular data-driven method is machine learning (ML). It is a well-known subfield of the field of artificial intelligence. It is the study of how to use computers to imitate human knowledge activities. This field also investigates the methods by which computers can improve themselves to acquire new knowledge and new skills, recognize previously acquired information, and continuously improve their performance and achievement [3]. This study is an implementation of ML in the domain of the medical field. This study uses machine learning and deep learning technique to detect IM.

Machine learning and deep learning provide new methods for predicting mortality risk and classifying risk variable. The study aims to apply ML to analyze data of several traits involving mothers during pregnancy that can lead to neonatal infant death. This study also creates a model based on a machine learning algorithm that was trained by utilizing data from both successful and unsuccessful deliveries in the past. Different ML techniques such as K-nearest neighborhood (KNN), logistic regression (LG), support vector machine (SVM), XgBoost, random forest classifier (RF), long short-term memory (LSTM), and convolutional neural network (CNN) to analyze IM to determine which risk variables are most strongly linked to IM. The accuracy of each model was compared to determine the most effective techniques to forecast newborn mortality. The dataset used to train the models were based on secondary information on Children's births and deaths in the city of São Paulo [4].

Many previous researches were done in the context of newborn child death. Purwanto *et al.* [5] worked on machine learning techniques such as autoregressive integrated moving average (ARIMA), Linear regression, and Neural networks, using mean absolute error and root square error. However, it heavily focuses on prediction rather than causal inference. Latif *et al.* [6] focus on the intra-die variation probe (IDVP) method of screening out infant mortality rate (IMR) due to system on-a-chip (SoC). They have claimed that IDVP is more suited compared to electronic tests. The scope of this study is very limited and only focuses on specific a method which is SoC device. Dereje *et al.* [7] use the technique of bagging and AdaBoost to discover possible threats and newborn mortality specifically for Ethiopia, with an accuracy of 94.39%. They have indicated that maternal health is the main reason behind infant mortality. Furthermore, in the study, Chowdhury *et al.* [8] aim to predict the health of the fetus or unborn child using a neural network. They claim that out of 100 fetal births, 9 have some sort of health issue. Although they discuss infant death but focus on fetal health. Another study that uses machine learning is by Hu *et al.* [9] records the vital signs of preterm infants in the neonatal intensive care unit (NICU) constantly and develops a non-invasive way based on ML techniques to anticipate the treatment given to them by the clinicians. However, there is a concern that the findings have limited applicability. Another study was conducted on 259 babies' vitals such as heart rate variability feature as input data to train their model. The study was conducted by León *et al.* [10], and used a recurrent neural network to discover late-onset sepsis (LOS) of preterm babies. The data was very limited in this study which was otherwise an impactful work. Another study that was conducted by Moreira *et al.* [11] determines that the weight of the fetus leads to serious issues during pregnancy. The data collected for the study was captured worldwide. They have used ML approaches, such as Random Forest, Decision Tree, Support Vector Machine, and Logistic Regression. Another study by Mfateneza *et al.* [12] analyzes data from the Rwanda Demographic and Health Survey to predict infant mortality using conventional methods, such as STATA version 13, and ML approaches, such as Random Forest, Decision Tree, Support Vector Machine, and Logistic Regression. Reidpath *et al.* [13] suggest that using measures of population health, such as disability-adjusted life expectancy (DALE), maybe more comprehensive than using IMR. However, the limitation of using broader measures of population health is that they may require more resources and could divert attention away from urgent initiatives. Brahma *et al.* [14] use ML to predict newborn and IM incidence in India and identify early warning indicators for high-risk populations. They used ML to identify early predictors of infant mortality and provide targeted interventions for high-risk populations. However, the data set is very limited in terms of scope and volume. Moreira *et al.* [15] aim to identify early predictors of newborn mortality using Artificial Intelligence (AI) and evaluate previous neonatal prediction studies. However, the disadvantage is that it only focuses on neonatal mortality and may not consider other important factors of population health. Brahma *et al.* [16] have worked with Machine Learning techniques to figure out the root cause of infant malnutrition. They have worked with various factors such as safe water, vaccines, vitamin A, ration cards, infant characteristics, mother health, and many more. According to their study, safe water is the deciding factor that should be focused on. They focused more on the solution rather than identifying the issue of predicting the issue to take proper action. Rahman *et al.* [17] use ML algorithms to evaluate potential factors linked to IM in Bangladesh. They used ML to identify potential factors linked to IM, which can provide targeted interventions for high-risk populations. However, they focused on Bangladesh and may not apply to other populations thus making the model very generic. Hajipour M *et al.* [18] identify significant predictors of infant mortality using various data mining algorithms, such as SVM, Artificial Neural Networks, KNN, RF, Naive Bayes, and AdaBoost. Although they tried various techniques it only focused on eight Iranian provinces and may not apply to other populations. Hahn *et al.* [19] compare the coding of race and ethnicity at birth and at death to compute IM rates in the United States. Anyhow it does not focus on predicting infant mortality or identifying early predictors of high-risk populations.

## 2. RESEARCH METHOD

In Figure 1, the methodology of this study is visualized. The main goal of this study is to find out the most optimal model to pre-determine the health of a child. To the most optimal model, the first data related to newborn babies were collected. The collected data was pre-processed with a series of steps that are described below. After the preprocessing nearmiss method was used on the processed data. Later the test and train data were split before training the Machine learning models. However, in the deep learning models the nearmiss method was not required. The confusion matrix for all the models were generated in order to compare the results.

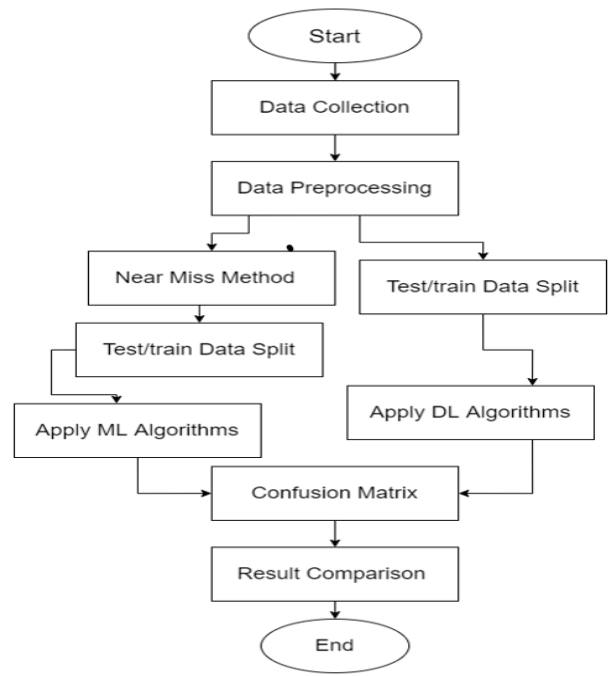

Figure 1. Flowchart of the methodology

### 2.1. Data collection

This dataset is based on secondary information on Children's births and deaths in the city of São Paulo between 2012 and 2018 [4]. There are 24 columns and 1,427,906 rows in the dataset. All the column descriptions are given in Table 1. In the data set, the first 23 columns are feature columns, and the final column named Death within 28 Days is the label column. Figure 2 describes some of the features of data set.

Heatmap helps to identify the correlation between each and every column of the dataset. It helps to better correlate the features and the label. For each value plot, there is a heatmap value that indicates different shades of the same color. Deep colors on the chart often reflect higher values than lighter shades. A completely different color can be used for a very different quality in the same way. The example below is a two-dimensional plot of values that are mapped to the indices and columns of the chart. The heat map of the dataset is given in Figure 3.

Table 1. Comparison between each model on basis of different accuracy measurements

| Algorithm name | Accuracy | Precision | Recall | F1-score |
|---|---|---|---|---|
| Logistic regression | .93 | .93 | .93 | .93 |
| KNN | .90 | .90 | .90 | .90 |
| Support vector classifier (SVC) | .87 | .87 | .97 | .87 |
| Xgboost classifier | .94 | .93 | .93 | .93 |
| Random forest classifier | .94 | .93 | .93 | .93 |
| CNN | .98 | 1.00 | .98 | .99 |
| LSTM | .99 | .99 | .99 | .99 |

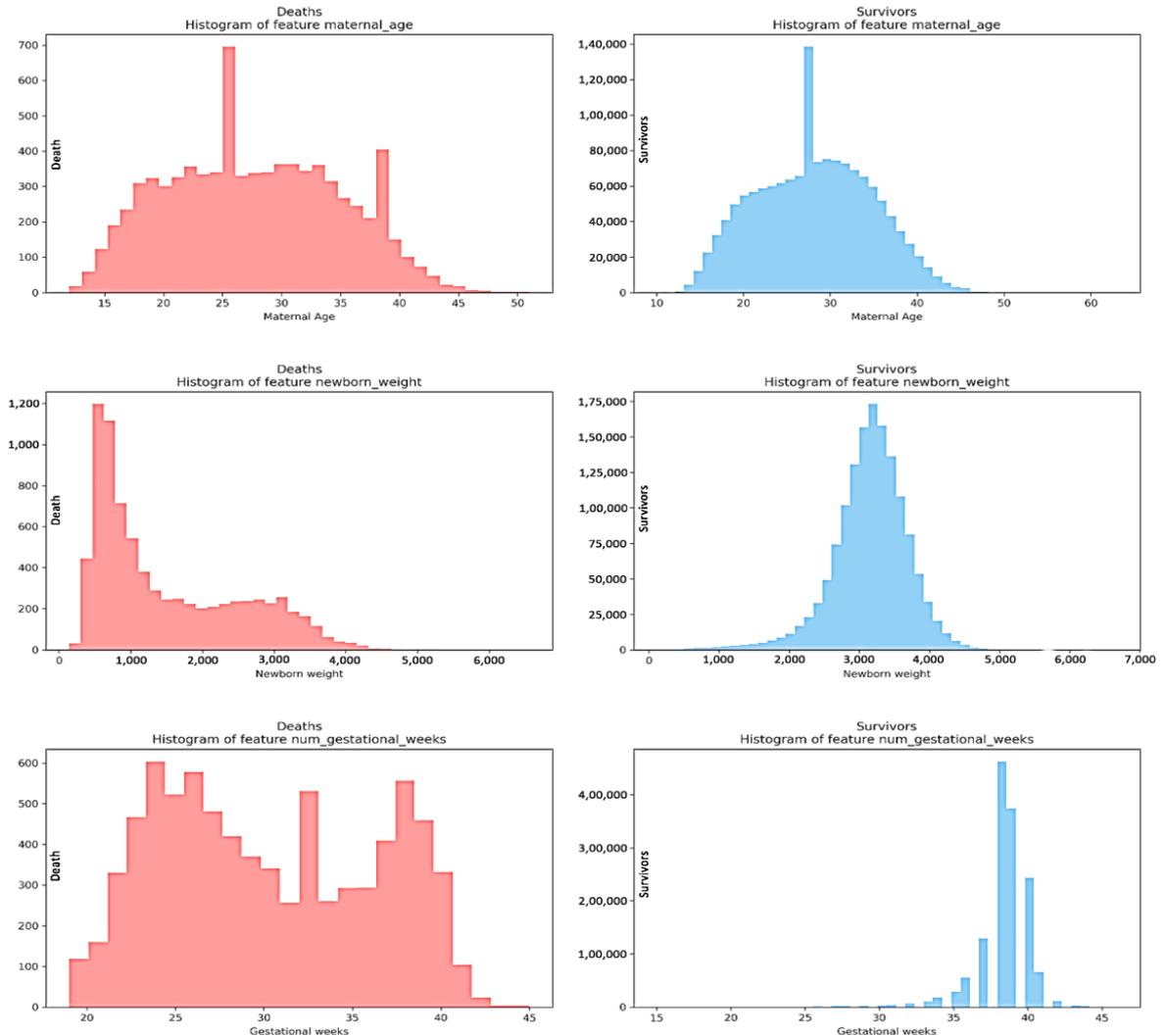

Figure 2. Value count of categorical features from the data set

**2.2. Data pre-processing**

From the data set, the null value and missing data were identified. This would lead to an anomaly in the trained models. To resolve the issue, the rows were deleted if a single missing value was found to be null. The missing values were eliminated using NumPy's dropNa method.

**2.3. Near miss method**

The analysis begins with the identification of the imbalance of the label column. From In Figures 4 and 5, it is evident that the count of label values where the cases being 0 is much higher compared to the outcome being 1. This causes further biasness and overfits of the models while training. To reduce the imbalance of data, the Near Miss technique was used. After doing under-sampling, the results are visualized in Figure 6.

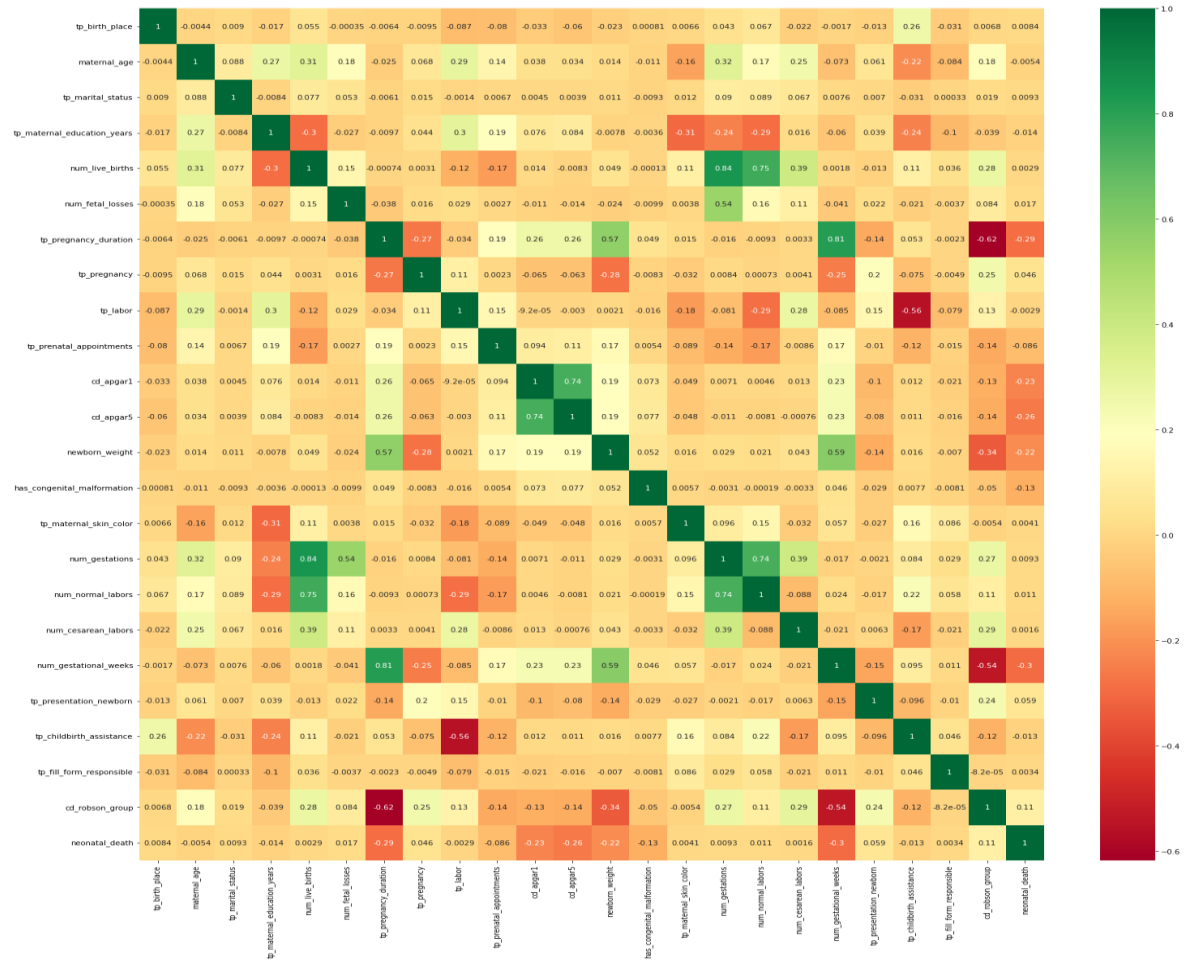

Figure 3. Correlation between each dataset column with heatmap

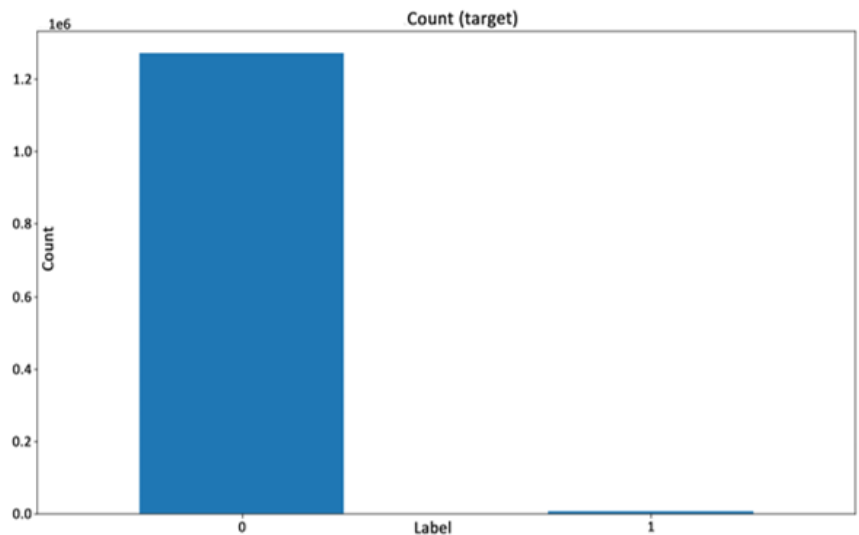

Figure 4. Count of label values before applying nearmiss

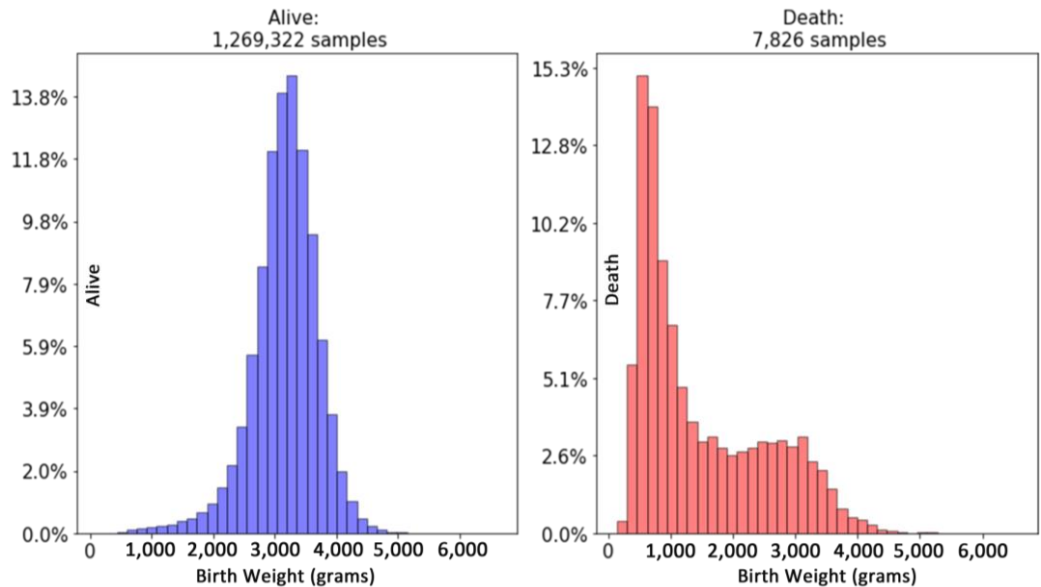

Figure 5. Analysis between alive and death in weight

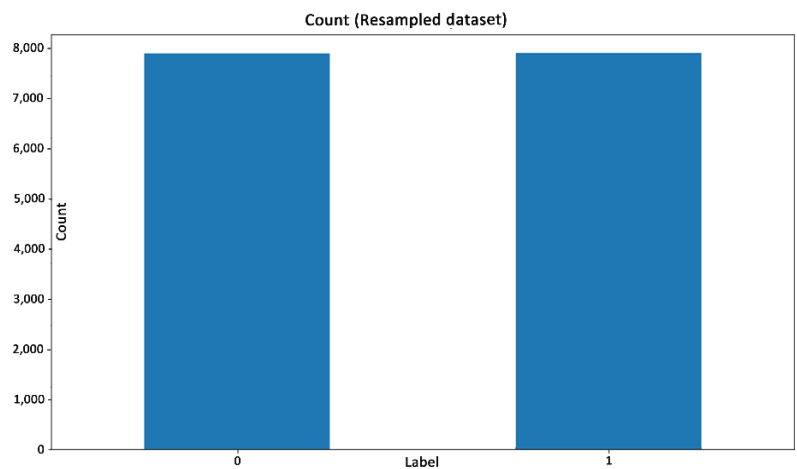

Figure 6. Count of label values after applying nearmiss

**2.4. Test/train data split**
The Feature and label parts of the dataset were divided. Then both the feature and label set are divided into two parts. This study maintains a 70:30 ratio to split the data, which means that 70% of the data is used to train ML models, and the rest 30% of the data is used for testing the trained, intelligent model to figure out the accuracy. To split the data set, the scikit-learn library's train test split method is used. The method randomly chooses the columns for test and training.

**2.5. Apply ML algorithms**
After splitting the data set, various machine-learning methods were used to train several different models. The selected methods to train the models were KNN, random forest classifier, support vector classifier (SVC), Logistic Regression, and XG Boost. To train, each model sklearn library was used.

**2.6. Apply deep learning algorithms**
The machine learning algorithm accuracies were quite good. However, to increase the accuracy of the model, we attempted deep learning methods as well. When implementing deep learning models, the near miss method was evaded because the model was performing quite well despite not applying to an under-sample. Thus, the model was trained with the complete data set. CNN and long LSTM algorithms were used. The accuracy of both approaches was better compared to the supervised algorithms. TensorFlow and Keras library was used to train the algorithms. The description of each algorithm is given below.

**3.6.1. Logistic regression (LR)**
Logistic regression (LR) is known as a statistical method that is comparable to linear regression, but logistic regression can be used for both categorical and continuous outcomes. To predict grouping LR uses a log odd ratio and iterative maximum likelihood method instead of the last square to fit the final model [20]. It is mostly used for prediction and classification problems.

**3.6.2. K-nearest neighbor (KNN)**
K-nearest neighbor (KNN) is a machine learning algorithm that uses a distance matrix namely Euclidean distance. It tries to discover the closest example in the training set and classifies points based on these neighborhood points [21]. It's mostly used for pattern recognition, recommendation engine & data mining.

**3.6.3. Support vector classifier**
Support vector classifier is often associated with SVM. It is typically used for classification tasks. Like SVM the SVC also searches for the optimal separating surface. SVC is outlined first for the linearly separable case [22]. It is mainly used for binary classification and potential extension to multi-class problems.

**3.6.4. Extreme gradient boosting (XgBoost)**
Gradient boosting is a regression algorithm that operates in a boosting-like fashion. Its primary objective is to approximate a function, denoted as $\hat{F}(x)$, which takes instances x as input and predicts their corresponding output values y. This approximation is achieved by minimizing the expected value of a specified loss function, denoted as $L(y, F(x))$. Gradient boosting constructs an additive approximation for $F*(x)$ by creating a weighted sum of functions. XGBoost, a technique built upon the principles of gradient boosting, stands out as a dependable and highly efficient machine learning algorithm [23].

**3.6.5. Random forest classifier**
It is an ensemble learning method. It is fast in operation and simpler to implement yet proven to be extremely effective in varieties of the domain. The general idea behind the algorithm is to construct much simple decision trees and vote across them in the classification stage [24].

**3.6.6. Convolutional neural network (CNN)**
CNN is a deep-learning algorithm with several layers. The first structure where the convolutional operation happens is known as the convolutional layer. Here the operation is done on the input data, and then the output is sent to the consequent layer. The second layer is known as the pooling layer. It combines the collection of the output of the previous layer into a single neuron. The next layer is a fully connected layer where each layer is connected to all the layers. There could be multiple layers before the output layer. The mechanism of CNN is picturized below in Figure 7 [23].

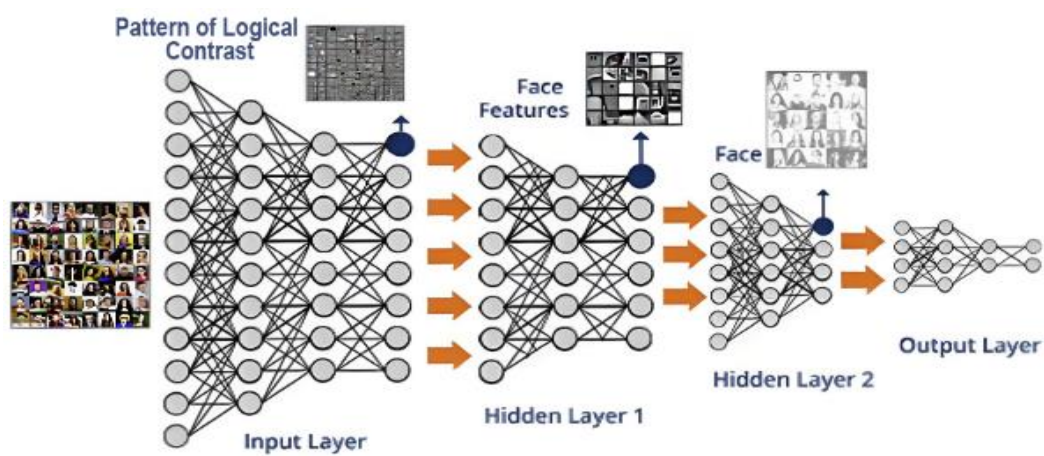

Figure 7. Convolutional neural network [25]

### 3.6.7. Long short-term memory (LSTM)

LSTM is a special kind of Recurrent Neural Network that can selectively remember patterns for a long duration time [25]. While it is a perfect choice for sequential data but it has been proven quite formidable in other aspects as well. The structure of LSTM is visualized in Figure 8.

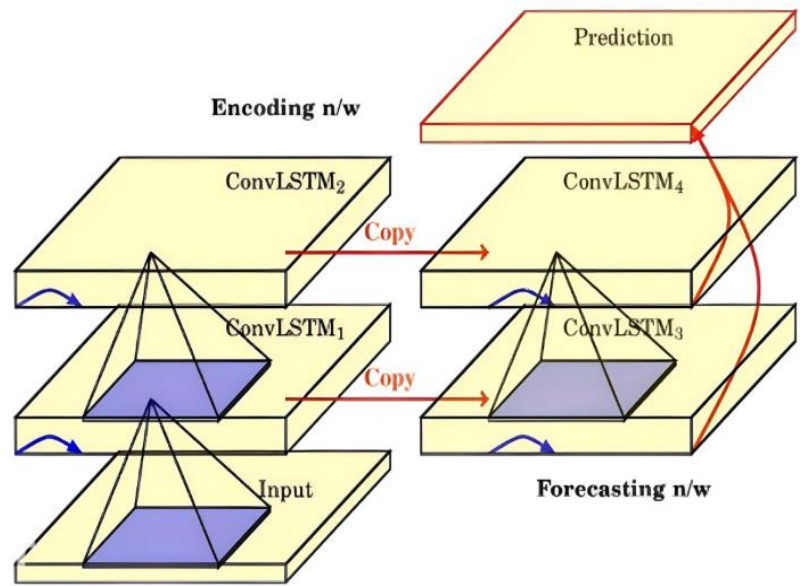

Figure 8. Structure of a convolutional long short-term memory [26]

## 3. RESULTS AND DISCUSSION

All the recent previous studies relating to or closely similar to our titular topic were reviewed above. Most of these studies had either of the following few limitations such as data limitation, limited scope of study, lack of diverse techniques, focus too much on the solution rather than identifying the root cause, etc. While the past studies [5], [7], [12], [14], [16], [18] focused on ML or developed predictive models have done very good work, we implemented many more ML models along with deep learning models ensuring better accuracy. Many other past works utilized a data-driven approach [9], [14], [19] but they had a very limited dataset or focused on a small demographic only. On the other hand, we focused on diverse datasets and large-scale datasets. However, in our conducted studies, we tried to work with a data set of globally recommended organization datasets and the quantity of the data set is very large. Additionally, we applied various machine learning and deep learning techniques to figure out the most optimal model that can predict the mortality of a newborn. We have not only come up with the most accuracy. Again, we also made sure that the data set was not biased thus we applied a sampling technique.

To compare the techniques, a confusion matrix was generated for each of the algorithms [26]. A confusion matrix provides an in-depth view of each model's accuracy [27]. Different categorization metrics were also generated, and these metrics can help you choose the most optimal model. A confusion matrix outputs four different outcomes such as true negative (TN), true positive (TP), false positive (FP), and false negative (FN) [28]. Each formula of the outcomes is defined below. The outcomes of the confusion matrix help to find various decision mechanisms such as accuracy, recall, precision, and F1 score [29], [30]. All these criteria help to identify how well the method is trained and how well the method will perform.

a) True Positive: A true positive outcome occurs when the model properly predicts the positive class. a true positive result (TPR) is calculated as TP/TP + FN [31].
b) False positive: A result that is incorrect to be a positive result. False positive rates are calculated as FP/FP + TN [32].
c) True negative: A result that is correct to be negative. True negative rates are calculated as TN/TN+FP [33].
d) False negative: A result that is incorrect to be a negative result. False negative rates are calculated as FN/FN+TP [33].

The accuracy of a model depends on all the previously mentioned aspects. Accuracy classification is defined by the data model as the exact expectation ratio of the total expectations generated by the classifier. Accuracy is calculated using. $(TP+TN)/(TP+FP+TN+FN)$.

**Precision:** Precision is a statistic that measures the amount of accurate positive predictions. Precision is calculated using $TP/(TP + FP)$ [34].

**Recall:** This quantity of precise good predictions out of all potential positive predictions is measured by the recall. Recall is calculated using $TP/(TP + F)$ [35].

**F1 Score:** The F-score, commonly identified as F1-score, is the accuracy of the model in a dataset. The equation of calculating F1 score is $(2 * precision * recall)/(precision + recal)$ [35].

The confusion matrix of each algorithm was generated using Sklean's confusion_matrix function. The results are given in Table 1 for each algorithm. For this study, seven algorithms were applied with the same data set. Out of all the machine learning algorithms, both the Random Forest classifier and XGboost perform the best with 94% accuracy. And out of the deep learning models, the LSTM was proven to be 99% accurate. The accuracy of each model was compared graphically in Figure 9.

This means that the LSTM model has the most accuracy out of all the experimented algorithms. Thus choosing this model to use it for real-life task would lead to the most accurate outcome. After inputting the data of a newborn children this model will predict if the child's health is at risk. Consequently, adequate precaution can be taken to ensure the baby's wellbeing.

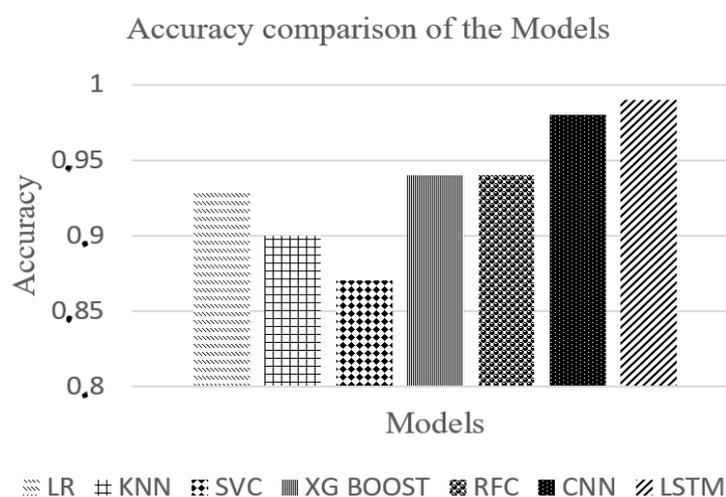

Figure 9. Accuracy comparison between different models

## 4. CONCLUSION

Data-driven decision-making is used in almost every aspect of modern times. This detailed study looks to analyze data based on a very critical issue for developing and underdeveloped countries. To create a robust model that can predict infant mortality rate, data sets were analyzed to reduce data imbalance, and later using the processed data, various predicting models were trained. Both deep learning and machine learning were used. Out of the trained models, the Random Forest classifier and XGboost were found to be the most accurate based on various criteria. Again, out of the deep learning model, the LSTM model was the most accurate. These intelligent models will help doctors and related people to be aware beforehand to be prepared about a newborn child as it can predict a child's mortality with up to 99% accuracy. The accuracy means that out of the 100 times input parameters are passed, the model will result in correct output 99 times. This will boost the early detection of endangered children. Early prediction of a situation leads to being more cautious about taking better care and support of the newborn. This can help the stakeholders to be better prepared for taking care of a newborn infant.

## BIOGRAPHIES OF AUTHORS

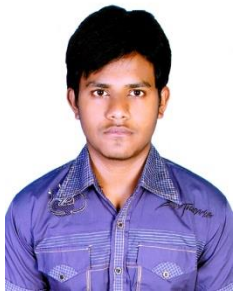

**Mohon Raihan** completed his B.Sc. in Computer Science and Engineering from American International University Bangladesh (AIUB). His research interest is Machine Learning, Data science, Software engineering. In addition to his academic pursuits, Mohon actively contributes to open-source Machine Learning projects and is passionate about knowledge sharing. He can be contacted at email: mohon.raihan1020@gmail.com.

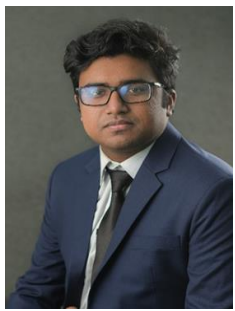

**Plabon Kumar Saha** completed his undergrad from American International University Bangladesh (AIUB). His research interest is Machine learning, Data science, Software engineering and IoT. He has 11 publications to his name and has over 27 citations. He can be contacted at email: pkumarsaha71@gmail.com.

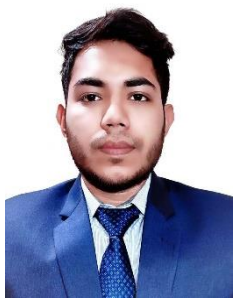

**Rajan Das Gupta** is an accomplished machine learning researcher specializing in AI. He holds a B.Sc. in Computer Science and Engineering from American International University-Bangladesh and an M.Sc. in Computer Science from Jahangirnagar University. His research focuses on innovative algorithms for deep learning and reinforcement learning. His research interest is Machine learning, Data science and IoT. He can be contacted at email: rajandasgupta.me@gmail.com

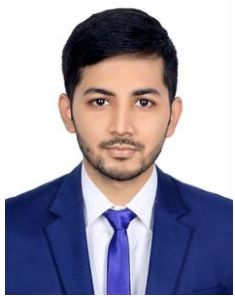

**A Z M Tahmidul Kabir** received his B.Sc. Eng. Degree in Electrical and Electronic Engineering from American International University-Bangladesh. He has authored or coauthored more than 15 publications: 15 conference proceedings and 5 journals, with 6 H-index and more than 100 citations. His research interests include Embedded systems, IoT, Robotics, and Autonomous Vehicles. He can be contacted at email: tahmidulkabir@gmail.com.

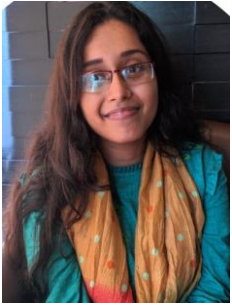

**Afia Anjum Tamanna** is a frontend Software Engineer with a keen interest in machine learning research. She is currently leading frontend web development at her own company and has extensive experience in designing user-friendly interfaces for web applications. Afia is passionate about exploring the possibilities of machine learning and actively engages with the community. She can be contacted at email: afiatamanna06@gmail.com.

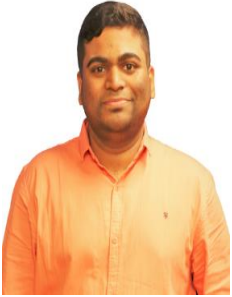

**Md. Harun-Ur-Rashid** is a Software Engineer at Giga Tech Limited-Bangladesh located in Dhaka, Bangladesh. Currently he is a Graduate student at United International University-Bangladesh located in Dhaka, Bangladesh. He is currently an aspiring deep learning enthusiast who aims to implement his skills for the betterment of public health, agriculture. He has one published paper on deep learning based in IEEE explore. He can be contacted at email: engharunurrashid97@gmail.com.

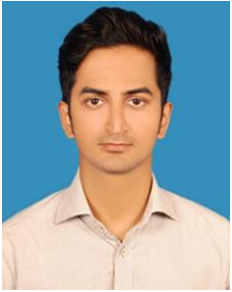

**Adnan Bin Abdus Salam** is a Computer Science and Engineering graduate from American International University-Bangladesh located in Dhaka, Bangladesh. He worked as a Mentor for Bangladesh Generation Parliament – A platform by UNICEF Bangladesh and Bangladesh Debate Federation to teach the children of Bangladesh the art of policymaking. He is currently an aspiring machine learning engineer who aims to implement his skills for the betterment of public health. He can be contacted at email: adnansalam007@gmail.com.

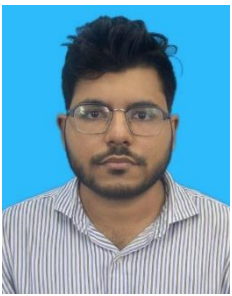

**Md Tanvir Anjum** has received his B.Sc. Computer Science and Software Engineering from American International University, Bangladesh. His research interests include software methodology and process, Human computer interaction, Machine learning, Statistical modeling and methods, product management process. He can be contacted at email: mdtanviranjum21@gmail.com.

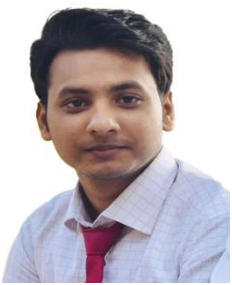

**A Z M Ahteshamul Kabir** has received his B.S.S & M.S.S in Information Science and Library Management from University of Rajshahi. He is currently pursuing his second Master's in Predictive Analytics at Curtin University. His research interests include Machine learning and Data science. He can be contacted at email: arath.kabir@gmail.com.